# Non-invasive Diabetes Detection using Gabor Filter: A Comparative Analysis of Different Cameras

Christina A. Garcia[1], Patricia Angela R. Abu[2], and Rosula S. J. Reyes[3]

**Abstract**   This paper compares and explores the performance of both mobile device camera and laptop camera as convenient tool for capturing images for non-invasive detection of Diabetes Mellitus (DM) using facial block texture features. Participants within age bracket 20 to 79 years old were chosen for the dataset. 12mp and 7mp mobile cameras, and a laptop camera were used to take the photo under normal lighting condition. Extracted facial blocks were classified using k-Nearest Neighbors (k-NN) and Support Vector Machine (SVM). 100 images were captured, preprocessed, filtered using Gabor, and iterated. Performance of the system was measured in terms of accuracy, specificity, and sensitivity. Best performance of 96.7% accuracy, 100% sensitivity, and 93% specificity were achieved from 12mp back camera using SVM with 100 images.

## 1 Introduction

In the emergence of coronavirus disease (COVID) pandemic, patients with existing chronic diseases have higher fatality rates. Considered among the most common chronic disease is diabetes. According to the Centers for Disease Control and Prevention, diabetes is one of the top three health problems of COVID-19 patients [1]. The International Diabetes Federation (IDM) Atlas 9th Edition 2019 estimates 4.2 million deaths due to diabetes and projects the total people worldwide with diabetes will rise to 700 million by 2045 [2].

C. A. Garcia (christina.garcia@obf.ateneo.edu)
P. A. R. Abu (pabu@ateneo.edu)
Department of Information Systems and Computer Science
Ateneo de Manila University, Loyola Heights, Quezon City, 1108, Philippines

R. SJ. Reyes (rsjreyes@ateneo.edu)
Department of Electronics, Computer, and Communications Engineering
Ateneo de Manila University, Loyola Heights, Quezon City, 1108, Philippines

Diabetes Mellitus is defined by World Health Organization (WHO) as a chronic, metabolic disease characterized by elevated levels of blood glucose. Insulin is the hormone that controls glucose in the blood. IDF estimates 374 million people at increased risk of developing type 2 diabetes in which the body is insulin deficient or insulin resistant [2]. With the less common type 1 diabetes, the body does not make insulin. Excessive glucose in the blood consequently damages the blood vessels, eyes, heart, or the nerves.

As reported by IDF, 3 out of 4 adults with diabetes are living in low-income and middle-income countries [2]. Affordable treatment including insulin is critical to the survival of diabetic patients. Similarly, DM detection and monitoring are important to avoid or address complications as early as possible. Blood tests are commonly done to detect diabetes. The common method is Fasting Plasma Glucose (FPG) test which takes up to 12 hours [4][5]. Traditional DM diagnostic methods such as glycated hemoglobin (A1C) test, FPG test, oral glucose tolerance test and 2-h plasma glucose test obtain blood sample from patients and do not guarantee immediate results. All four techniques are invasive causing pain and discomfort [6].

In the surge of the COVID-19 pandemic, patients with diabetes constitute a vulnerable population [7]. With the alarming rate of increasing DM cases in Western Pacific, a significant need for an affordable and accessible detection method for diabetes rises. In 2019, 1 every 2 DM cases were undiagnosed that equivalent to 232 million people. More than half of DM cases in WP remain undiagnosed and patients are at a higher risk of developing complications [2].

Painless non-invasive technique to detect DM is a promising technology reducing the hospital visit expenses and time. Early diagnosis and disease intervention for pre-diabetes are also important. In the recent years, researches on DM classification have surfaced with analysis focused on using facial blocks [8]. However, the restrictive set-up for image capture lengthens the procedure. With few literatures available, DM detection using facial texture features has not been much explored [6].

This study aims to compare and explore the performance of different cameras as convenient tool for image capture for non-invasive diabetes detection.

- To use three different cameras to build a dataset of facial images captured within natural lighting condition.
- To implement facial texture extraction using varied Gabor filter and classify different facial features using Support Vector Machine (SVM) and k-Nearest Neighbors (k-NN).
- To iterate data and compare the performance of different cameras in terms of accuracy, sensitivity and specificity.

## 2 Review of Related Literature

### *2.1 Image Acquisition and Dataset*

Reference [9] used a particular camera device for image acquisition which has been adopted by the majority of the researches on DM detection based on facial features. Similarly, the dataset of 426 images with 284 DM samples and 142 Healthy samples has been used in other studies.

A total of 40 images comprised the dataset of [12] with 20 DM samples and 20 healthy samples. The common ratio maintained for training and testing was 70:30 [6][12].

References [6][8][9] used a black box as an image capture device with a centered 3-CCD camera and lamp on each side. Meanwhile, [11] used a less restrictive set-up for image capture using a mobile device camera placed 30cm in front of each subject under normal lighting condition. The mobile camera used for capturing facial images has optical image stabilization.

### *2.2 Facial Blocks and Image Pre-processing*

References [4][8][11] extracted four blocks from the captured facial images after pre-processing. The facial blocks represent the main regions of the face according to Traditional Chinese Medicine (TCM). Block F represented the forehead while Block N represents the bridge of the nose. Blocks R and L were taken below the right and left eyes respectively were analyzed for DM detection.

Fig. 2.1 shows the extracted facial blocks from the original image of the participant from the dataset of [11]. Raw captured images include non-facial regions, so image cropping was performed. Image resizing was also applied. According to [13], pre-processing of the facial images is important in order to better discriminate the features to be obtained.

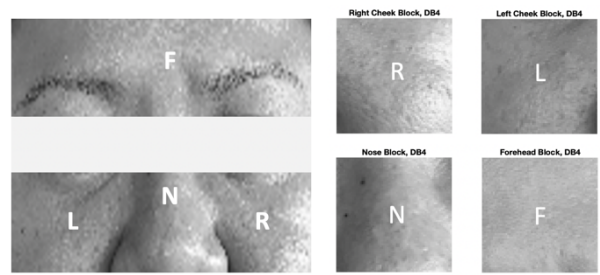

**Fig. 2.1** Labeled extracted facial blocks from the original image.

## *2.3 Facial Texture Feature Extraction*

Gabor filter was used by [11][14] to extract texture features from each facial block representing the sample image. Each block was described by texture value. A custom-sized 2-D Gabor filter bank comprised of 40 filters was generated with five scales and eight orientation combination. A column vector with one texture value for each filter as calculated from the texture features. The computed mean vector was assigned the texture value of the facial block. A response which was the texture value or mean of all pixels was produced from the convolution of the filter with facial blocks. The final texture value of a facial block was calculated by taking the mean of all forty texture values.

Reference [12] implemented Gray Level Co-occurrence Matrix (GLCM) on 40 images dataset to study extensively the effects of texture features on detecting DM. Contrast, correlation, energy, Haralick's features and homogeneity were extracted. On the other hand, four texture feature families were implemented to compare the effects of different texture feature extractors in detecting DM by [6][8]. Statistical and signal processing base texture feature families gave better outcomes in detecting DM according to their experiments. Statistical feature included GLCM, Image Gray-Scale Histogram, and Local Binary Pattern while signal based included Gaussian, Gabor Filters, and Steerable.

## 3 Methodology

Non-invasive DM detection system compared and used 2 sets of facial block features in classification, and 4 different methods for feature extraction and selection. Fig. 3.1 shows the general framework of the implemented system.

**Fig. 3.1** Implemented non-invasive DM detection system

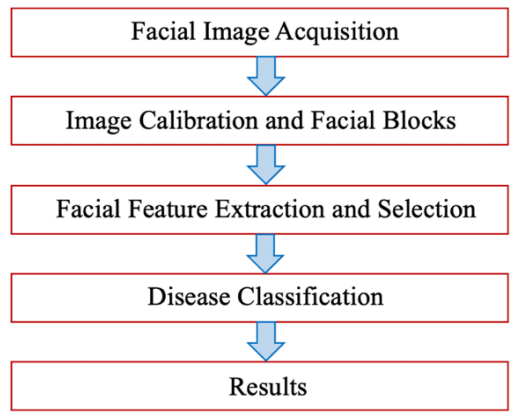

Diabetes Mellitus was detected through the analysis of texture features of three facial blocks from each of the images in the dataset. Training and testing sets of the database have undergone the same stages for classification.

## 3.1 Image Acquisition and Dataset

Listed below (Table 3.1) are the demographic details of the 50 diabetic participants from the different provincial locations. Similarly, 50 participants from the same provinces ages 20 to 69 years old were photographed for the healthy dataset. The ratio of 70:30 for training and testing implemented in the previous studies was maintained.

**Table 3.1** Diabetic Dataset with 50 participants

| Residency  | Female | Male | Age     | Years diagnosed | Type of DM |
|------------|--------|------|---------|-----------------|------------|
| No. 1 City | 11     | 12   | 20 - 70 | 1 - 20          | Type 2     |
| No. 2 City | 3      | 6    | 30 - 59 | 1 - 10          | Type 2     |
| No. 3 City | 3      | 2    | 30 - 70 | 8 - 40          | Type 2     |
| No. 4 City | 7      | 5    | 52 - 87 | 2 - 34          | Type 1-2   |
| No. 5 City | 1      |      | 40 - 49 | 5               | Type 2     |

For each subject, the 12mp back camera and 7mp front camera of an iPhone 7 with image stabilization were used to take the photo of each participant. A laptop camera of 720p MacBook Air was also used for comparison. Each device was placed approximately 30 cm in front of the face under normal lighting condition.

## 3.2 Facial Regions and Image Pre-processing

The blocks obtained from each image of the dataset were all resized to 64 x 64, specifically forehead, nose, left cheek, and right cheek. Each image was converted into grayscale from original RGB (truecolor) for the Gabor filter. All facial block images were made uniform in size prior feature extraction stage. Image pre-processing techniques; cropping, resizing, and color conversion were performed to better discriminate the texture features.

## 3.3 Gabor Filter and Texture Feature Extraction

In this study, Gabor filter was characterized with varying lambda $\lambda$ (wavelength) and theta values $\theta$ (orientation) at different degrees of interval to find the best parameters for texture extraction of Filipino participants. Generally, 2-D Gabor filter is a Gaussian kernel multiplied by a sinusoid and has parameters namely: wavelength, orientation, aspect ratio, and bandwidth. The wavelength controls the width of the strips of the function while the orientation governs the orientation of the Gabor envelope.

In the equation, $G_k(x,y)$ represents the filter while $im(x,y)$ is the block of facial image. * is for 2-D convolution (Eq. 3.3).

$$R_k(x,y) = G_k(x,y) * im(x,y) \tag{3.3}$$

One facial sample represents the combination of facial blocks, each having texture values. The texture value of each response is the mean of all of its pixels. The final texture value of a facial block can be calculated by taking the mean of all texture values. For this study, two different sets of texture features were extracted from each facial block as predictors for classification.

Four methods with varying Gabor filters and texture features for facial block feature extraction were compared. Two different sets of texture features were extracted from each facial block as predictors for classification. Fig. 3.2 shows the framework for choosing Gabor filters which are uniform in all three facial blocks for single texture feature (method 1) and multiple texture features (method 2). Fig. 3.3 shows the framework implemented for choosing Gabor filters unique per facial block for single texture feature (method 3) and multiple texture features (method 4).

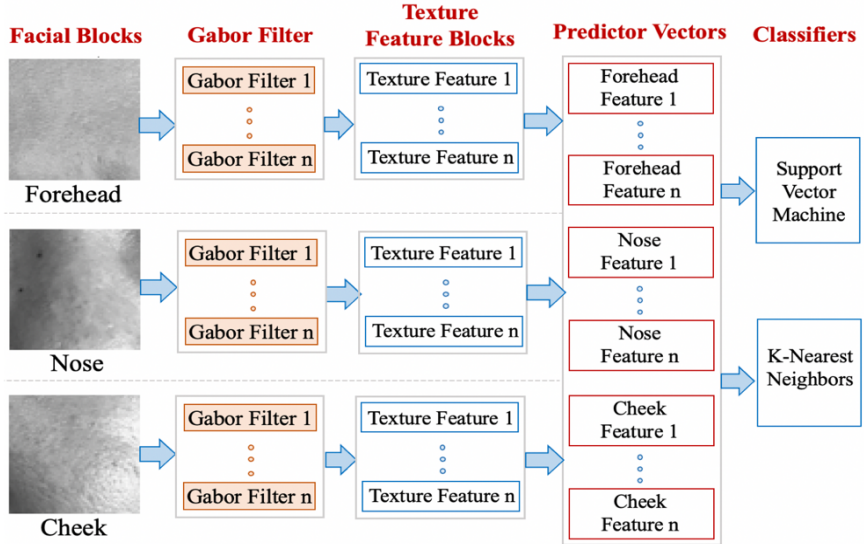

**Fig. 3.2** Framework for choosing Gabor filters uniform in all three facial blocks for single texture feature (method 1) and multiple texture features (method 2)

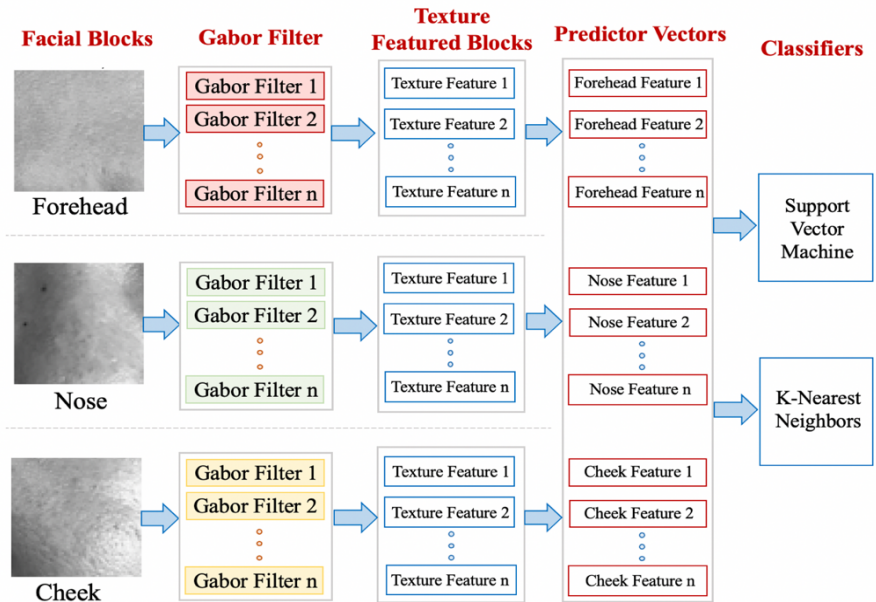

**Fig. 3.3** Framework for choosing Gabor filters unique per facial blocks for single texture feature (method 3) and multiple texture features (method 4)

The final Gabor filters implemented in this study has a total of 40 GFs in the filter bank with combination of 5 wavelengths and 8 orientations, starting from 0 to 360 degrees.

For method 1 with uniform Gabor filter per facial block, the mean texture vector was calculated similar to previous studies and combined vectors from each block was used as predictor to the classifiers. Similarly, for method 2 with uniform Gabor filters per facial block, a total of 6 texture features were obtained from each block. The predictor vectors were based on the mean, variance, kurtosis, std, entropy, and skewness. Every facial sample was represented by the combination of texture features from each of the three facial blocks.

Different Gabor filters were used for each facial block in methods 3 and 4, as opposed to the uniform Gabor filter for all three blocks in methods 1 and 2. For methods 3 and 4, different Gabor filters were characterized with varying lambda λ (wavelength) and theta values θ (orientation) at different degrees of interval per facial block. Classification performance of the system was evaluated separately using mean texture vector from each block. The best Gabor that resulted to high performance per block was chosen for the respective block.

For method 3, the mean texture vector was calculated similar to method 1 and combined vectors from each block was used as predictor to the classifiers. For method 4, a total of 6 texture features were obtained from each block similar to method 2. Every sample was represented by the combination of texture features from each of the three facial blocks.

## 3.4 Classification and Performance Parameters

The extracted texture features from each facial block are represented by predictor vectors $v_F$, $v_C$, and $v_N$ for forehead, cheek, and nose respectively. The resulting 3-D vector is a texture vector $v$ which is used as predictor to the classification algorithms SVM and k-NN. Four iterations of the images in the dataset were done to evaluate performance. Performance was measured in terms of accuracy, specificity and sensitivity.

# 4 Results and Discussion

Based on the summary of best performance of each camera per iteration (Table 4.1), it is notable that SVM performed better than k-NN as classifier for all three cameras. Also, method 2 with uniform Gabor filters used per block performed best among the 4 methods implemented for feature extraction. Increasing the predictors and dataset increases the performance.

**Table 4.1** Best Performance of 3 Cameras per Iteration

| Data | Camera/method | Accuracy | Sensitivity | Specificity |
|---|---|---|---|---|
| 100 | 12mp, m2, SVM | 96.7% | 100% | 93% |
| 100 | 7mp, m2, SVM | 90% | 87% | 93% |
| 100 | 720p, m2, SVM | 86.7% | 93% | 80% |
| 80 | 12mp, m2, SVM | 95.8% | 92% | 100% |
| 80 | 7mp, m2, SVM | 83.3% | 75% | 92% |
| 80 | 720p, m2, SVM | 87.5% | 92% | 83% |
| 60 | 12mp, m2, SVM | 94.4% | 100% | 89% |
| 60 | 7mp, m2, SVM | 83.3% | 89% | 78% |
| 60 | 720p, m2, SVM | 83.3% | 67% | 100% |
| 40 | 12mp, m2, SVM | 91.7% | 100% | 83% |
| 40 | 7mp, m2, SVM | 75% | 67% | 83% |
| 40 | 720p, m1, SVM | 83.3% | 67% | 100% |

Previous similar studies present only 2 to 6 cases which uses the same dataset or images from one high end camera, one to two feature extraction methods, and one to two classification algorithms to evaluate a non-iterated dataset. This research presented a comparison of 96 cases with the novelties of using new dataset captured from lower cost device, comparison of 3 different cameras, comparison of 4 texture feature extraction methods, and performance comparison at iterated values of dataset. Among the 96 cases tested, the 12mp mobile camera showed the highest performance per iteration. Fig. 4.1 shows the Area Under Curvature (AUC) ROC curve of 12mp resulting to 96.7% accuracy, 100% sensitivity and 93% specificity.

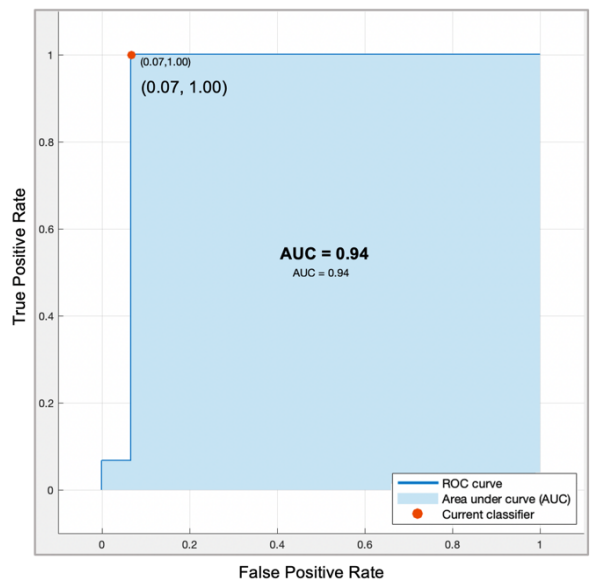

**Fig. 4.1** ROC of method 2, SVM for 12mp camera with 100 datasets

Based on the performance comparison (Table 4.1), out of the 96 cases studied, it can be observed that the resolution of the camera used for image capture, number of images in the dataset and texture features obtained from facial blocks affect the performance of the system. Among the 4 texture feature extraction methods, method 2 using the same Gabor filters for all 3 facial blocks with 6 texture predictors achieved the best accuracy for all dataset iterations. Also, the highest accuracy was attained with dataset of 100 images. As it is generally, increasing number of training images contribute to better system performance. Lastly, higher camera resolution contributes to increase in system performance. Between the 3 cameras used for data capture, the back camera with 12 megapixels achieved the best performance.

## 5 Conclusion and Recommendation

References [6][11][12][14] extensively reviewed and implemented various algorithms and techniques to detect DM. The best performance of achieved was compared with previous research works using non-invasive diabetes detection system (Table 4.2). The best performance was attained from the 12mp back camera with six texture features extracted using the same Gabor filters per facial block in method 2 with 100 images dataset.

**Table 4.2** Performance Comparison with Previous DM Detection Methods

| Study | Extractor | Classifier | Accuracy | Sensitivity | Specificity |
|---|---|---|---|---|---|
| Method 2, 12mp | Gabor | SVM | 96.7% | 100% | 93% |
| 2019 [11] | Gabor | SVM | 90% | 93% | 93% |
| 2017 [12] | GLCM | SVM | 91.67% | 100% | 83.33% |
| 2017 [6] | IGH | SVM | 99.02% | 99.64% | 98.26% |
| | Gabor | k-NN | 96.18% | 98.66% | 93.12% |
| 2014 [14] | Gabor | SVM | 99.82% | 99.64% | 100% |
| | | k-NN | 99.82% | 99.64% | 100% |

To improve the system performance, the researchers recommend further studies on factors that can improve the pre-processing stage. These factors may include addressing varied illumination, varying distance of camera from the face, and comparison of different image sizing of the facial blocks. Performance comparison of different signal processing base texture feature family aside from Gabor filter is also recommended. Studying the types of Diabetes and specific age range as contributing factors to performance aside from number of images captured is suggested. Trying low-cost devices for image capture will address equitability to health care.

**References**

1. Journal of Medical Emergency Services (2020), https://jems.com/2020/06/16/coronavirus-death-rate-is-higher-for-those-with-chronic-illnesses/. Accessed 10 October 2020
2. International Diabetes Federation (IDF Atlas. 9th ed., 2020), https://www.idf.org/aboutdiabetes/what-is-diabetes/facts-figues.html. Accessed 23 October 2020
3. World Bank collection of development indicators (World Bank, 2020), https://tradingeconomics.com/philippines/diabetes-prevalence-percent-of-population-ages-20-to-79-wb-data.html/. Accessed 29 October 2020
4. B. Zhang, P. Zhang, A study of diabetes mellitus detection using Sparse representation algorithms with facial block color features. IEEE International Conference in Signal and Image Processing (2016)
5. S. G. Thipe, K. R. Shirgire, P. A. Parab, U. Bhat, Diabetes mellitus detection based on facial block texture features using the Gabor filter. International Journal of Innovative Research in Science and Engineering, vol. 3, issue 4 (2017)
6. T. Shu, B. Zhang, Y. Y Tang, An extensive analysis of various texture feature extractors to detect diabetes mellitus using facial specific regions. Elsevier Journal in Computers in Biology and Medicine (2017)
7. A. E. Arcellana, C. Jimeno, Challenges and opportunities for diabetes care in the Philippines in the time of the COVID-19 pandemic. Journal of the ASEAN Federation of Endocrine Societies, vol. 35, no. 1 (2020), p. 55-57
8. S. N. Padawale, B. D. Jadhav, Non-invasive diabetes mellitus detection based on texture and color features of facial block. International Conference on Automatic Control and Dynamic Optimization Techniques (2016)

[9] B. Zhang, B. V. K. Vijaya Kumar, D. Zhang, Non-invasive diabetes mellitus detection using facial block color with a Sparse representation classifier. IEEE Transactions on Biomedical Engineering, vol. 61, no. 4 (2014)

[10] Z. Bing, W. Hongcai, Basic Theories of Traditional Chinese Medicine (2010)

[11] C. Garcia, R. SJ. Reyes, P. A. R. Abu, Non-invasive diabetes detection using facial texture features captured in a less restrictive environment. International Journal of Recent Technology and Engineering, vol. 8 (2019)

[12] S. Pavana, K, Shailaja, Diabetes mellitus detection based on facial texture feature using the GLCM. International Research Journal of Engineering and Technology, vol. 4 (2017)

[13] A. Sajjanhar, A. A. Mohammed, Face classification using color information. Information Journal (2017)

[14] S. Ting, B. Zhang, Diabetes mellitus detection based on facial block texture features using the Gabor filter. IEEE 17th International Conference on Computational Science and Engineering (2014)